\documentclass[10pt,twocolumn,letterpaper]{article}

\usepackage[pagenumbers]{cvpr} 
\usepackage{multirow}
\usepackage[table]{xcolor}
\usepackage{pifont}
\usepackage{algorithm}
\usepackage{algorithmic}
\usepackage{makecell}

\definecolor{cvprblue}{rgb}{0.21,0.49,0.74}
\usepackage[pagebackref,breaklinks,colorlinks,allcolors=cvprblue]{hyperref}

\title{Is Prompt Selection Necessary for Task-Free Online Continual Learning?}

\author{
Seoyoung Park \quad Haemin Lee \quad Hankook Lee\\
Sungkyunkwan University\\
{\tt\small seoyoungp@skku.edu \quad haemin.lee@skku.edu \quad hankook.lee@skku.edu}
}

\begin{document}
\maketitle
\begin{abstract}
Task-free online continual learning has recently emerged as a realistic paradigm for addressing continual learning in dynamic, real-world environments, where data arrive in a non-stationary stream without clear task boundaries and can only be observed once. To consider such challenging scenarios, many recent approaches have employed prompt selection, an adaptive strategy that selects prompts from a pool based on input signals. However, we observe that such selection strategies often fail to select appropriate prompts, yielding suboptimal results despite additional training of key parameters. Motivated by this observation, we propose a simple yet effective \textbf{SinglePrompt} that eliminates the need for prompt selection and focuses on classifier optimization. Specifically, we simply (\textit{i}) inject a single prompt into each self-attention block, (\textit{ii}) employ a cosine similarity-based logit design to alleviate the forgetting effect inherent in the classifier weights, and (\textit{iii}) mask logits for unexposed classes in the current minibatch. With this simple task-free design, our framework achieves state-of-the-art performance across various online continual learning benchmarks. Source code is available at \href{https://github.com/efficient-learning-lab/SinglePrompt}{https://github.com/efficient-learning-lab/SinglePrompt}.
\end{abstract}    
\section{Introduction}
\label{sec:intro}

\emph{Continual learning} \citep{chen2018lifelong} aims to train models on sequentially arriving data while mitigating catastrophic forgetting, which refers to the tendency of models to lose previously acquired knowledge due to the inability to revisit past data. Although many studies have explored continual learning in recent years~\citep{zhou2024continual, zhou2024class}, most existing work has focused on \emph{task-based offline} scenarios, where task boundaries are explicitly defined and the model is allowed to be trained sufficiently on the entire dataset for each task. However, these assumptions rarely hold in real-world scenarios. In practice, data typically arrive continuously without explicit task boundaries (e.g., task IDs), and each example can be observed only once for training. To better reflect such constraints, \emph{task-free online continual learning (OCL)} has emerged as a more realistic and challenging scenario, where the model must adapt in an online manner without access to task information.

Many studies in online continual learning have adopted parameter-efficient fine-tuning (PEFT) techniques, e.g., LoRA~\citep{hu2022lora, zanella2024low} and prompt tuning~\citep{lester-etal-2021-power, jia2022vpt}, to enable fast adaptation with minimal learnable parameters. Among them, prompt tuning has gained much attention due to its ability to adapt models injecting learnable tokens into a frozen pretrained models. To better handle diverse and evolving input distributions in continual learning, \emph{prompt selection} strategies have been considered in the recent literature~\citep{wang2022dualprompt,moon2023online,kang2025advancing}. This approach dynamically chooses the most relevant prompt for a given input by comparing the similarity between a query based on the input context and a set of learnable prompt keys. Notably, these prompt selection methods have demonstrated strong empirical performance across various OCL benchmarks.

\begin{figure*}[t]
    \centering
    \includegraphics[width=0.95\linewidth]{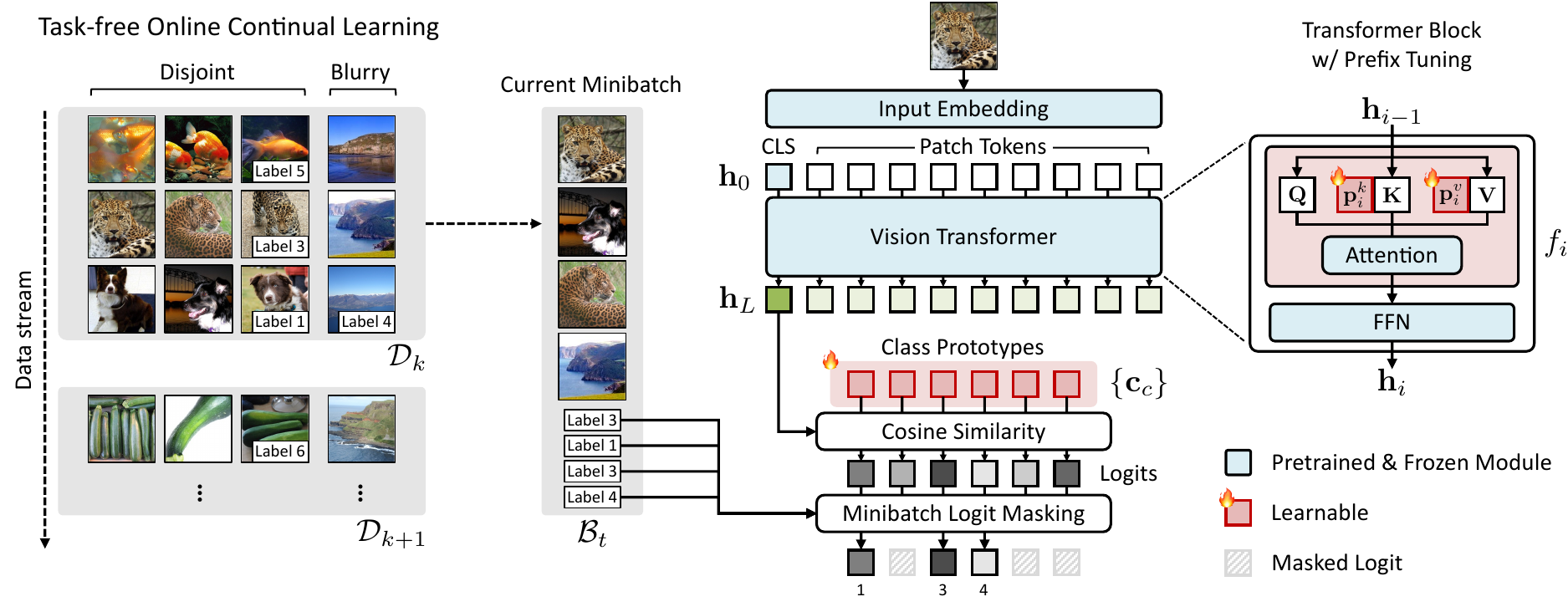}
    \vspace{-0.1in}
    \caption{An overview of the proposed \textbf{SinglePrompt}. When minibatch $\mathcal{B}_t=\{(\mathbf{x}_t^{(i)}, y_t^{(i)})\}_{i=1}^{B_t}$ is provided, it passes though a pretrained Vision Transformer encoder. At the $i$-th self-attention block $f_i$, the input sequence $\mathbf{h}_{i-1}$ is given, and during the attention operation the learnable prompts $\mathbf{p}_i^k$ and $\mathbf{p}_i^v$ are prepended to the key and value, respectively. Only the class token from the encoder's output sequence is used as the final representation. The cosine similarity between this representation and the class prototypes is computed and used as the logit values for prediction. The logit values corresponding to labels not exposed in the current minibatch are masked out before computing the cross-entropy loss.}
    \label{fig:main_fig}
\end{figure*}

Despite the strong performance of the recent prompt selection methods~\citep{wang2022learning, moon2023online, kang2025advancing}, interestingly, we observe that they often fail to select appropriate prompts. Specifically, under task-free continual learning scenarios, prompt selection tends to be largely uncorrelated with input data. Although the use of multiple prompts is based on the assumption that inputs from different labels will attend to different prompts, our experiments reveal that prompts selection is highly irregular with respect to input labels, with little evidence of clear separation among the prompts. Also, one experiment shows that prompt utilization is limited, with only a subset of the defined prompts being selected during inference. This challenge of ineffective prompt selection is not limited to task-free setting. In task-based continual learning, where task IDs are available during training and task-specific prompts can be trained, models~\citep{wang2022dualprompt, roy_2024_CVPR} still struggle to select appropriate prompts at the test time. For example, task-specific prompts are expected to be selected for inputs from the corresponding task, we however find that only about half of the prompts selected during evaluation match those assigned during training. 
This misalignment raises fundamental questions about the effectiveness of prompt selection.

Motivated by these findings, we introduce a simple yet effective framework, coined \textbf{SinglePrompt}, for task-free online continual learning. 
Instead of relying on potentially suboptimal prompt selection among multiple prompts, our framework simply uses a single prompt \emph{without prompt selection}. To be specific, we (\textit{i}) inject a single prompt per self-attention layer into a pretrained Vision Transformer~\citep{dosovitskiy2021vit}, (\textit{ii}) compute classification logits using cosine similarity to mitigate forgetting caused by the large norms of class prototypes, and (\textit{iii}) ignore classes not presented in the current minibatch by masking their logits. Our overall framework is illustrated in Figure~\ref{fig:main_fig}. This simple design enables our framework to require significantly fewer learnable parameters than existing baselines while achieving state-of-the-art performance across a wide range of benchmarks. We validate its effectiveness under the representative task-free online continual learning setting, Si-Blurry~\citep{moon2023online}, with three datasets: CIFAR100~\citep{krizhevsky2009learning}, Tiny ImageNet~\citep{le2015tiny} and ImageNet-R~\citep{hendrycks2021many}. For example, on CIFAR100, our method outperforms the prior state-of-the-art~\citep{kang2025advancing} by 6.55\% in $A_\text{last}$ while using approximately 60\% fewer learnable parameters. Our method also demonstrates strong performance across various task distribution shift scenarios, from the single-task setting to the fully disjoint (i.e., no overlap between tasks) setting, indicating its robustness to varying task conditions.

Overall, our work presents a new perspective on prompt-based adaptation in continual learning: rather than focusing on selecting the right prompt, a single well-designed prompt with an appropriate classifier can serve as a strong alternative. We hope this finding encourages the development of simpler, more robust approaches for task-free online continual learning.

\section{Preliminaries}
\label{sec:preliminaries}

\subsection{Setup: Task-free Online Continual Learning} 

In this paper, we mainly consider \emph{online continual learning (OCL)} scenarios, where data arrive in a streaming fashion and each example can be observed only once. An OCL model must learn continuously from a non-stationary data stream in an online manner, while maintaining performance on previously seen data distributions. It is particularly challenging because data distributions (often referred to as \emph{tasks}) in the stream can shift abruptly or gradually over time, and the model must mitigate catastrophic forgetting without access to previously observed data.

In general, OCL scenarios can be categorized into task-based and task-free settings. \emph{Task-based OCL} assumes that task boundaries are known or explicitly provided. Each incoming data point is associated with a task ID, which can be used to organize memory or modulate the behavior of the model. In contrast, \emph{task-free OCL} makes no assumption about task boundaries. Therefore, the model must adapt continuously without any external signal indicating distribution shifts. This setting better reflects real-world deployment scenarios, but is more challenging due to the absence of explicit task boundaries. Accordingly, we primarily focus on the task-free OCL setting, which we formally define below.

Let $(\mathcal{B}_1,\mathcal{B}_2,\ldots,\mathcal{B}_T)$ denote a sequence of minibatches arriving sequentially over time, where each minibatch $\mathcal{B}_t=\{(\mathbf{x}_t^{(i)}, y_t^{(i)})\}_{i=1}^{B_t}$ sampled from the corresponding task $\mathcal{D}_k$. Following the recent literature~\citep{wei2024online, kang2025advancing}, we use the Si-Blurry~\citep{moon2023online} scenario, in which task boundaries are blurry (i.e., tasks are gradually shifted over time), making it well aligned with the task-free OCL setting. Specifically, the $k$-th task $\mathcal{D}_k$ is associated with a label space $\mathcal{Y}_k\subseteq\{1,\ldots,C\}$ where $C$ is the total number of classes. Each sample $(\mathbf{x}_t,y_t)\in\mathcal{B}_t$ lies in $\mathcal{X}\times\mathcal{Y}_k$, where $\mathcal{X}$ is the image space. To induce blurry task transitions, $\mathcal{Y}_k$ is sampled from a mixture of disjoint and overlapping (i.e., blurry) class subsets. Importantly, due to the task-free nature, the model has no access to the task-specific label space $\mathcal{Y}_k$ or any information about task boundaries. For evaluation, we compute the average accuracy over all previously observed classes $\bigcup_{s=1}^k\mathcal{Y}_s$.
\begin{figure*}[t]
     \centering
     \begin{subfigure}{0.33\linewidth}
         \centering
         \includegraphics[width=\linewidth]{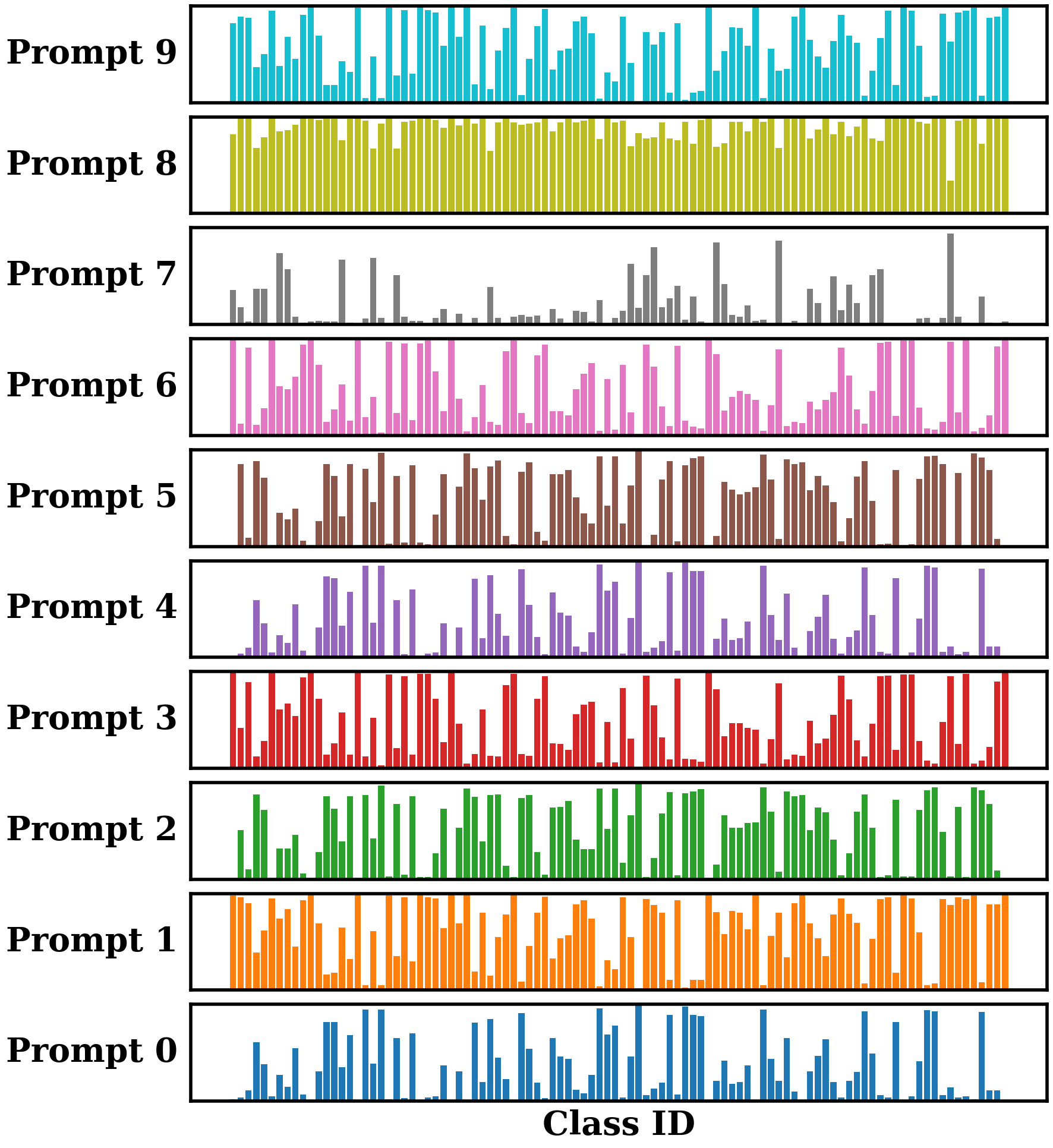}
         \caption{L2P~\citep{wang2022learning}}
         \label{fig:l2p_selection_cnt}
     \end{subfigure}
     \hfill
     \begin{subfigure}{0.33\linewidth}
         \centering
         \includegraphics[width=\linewidth]{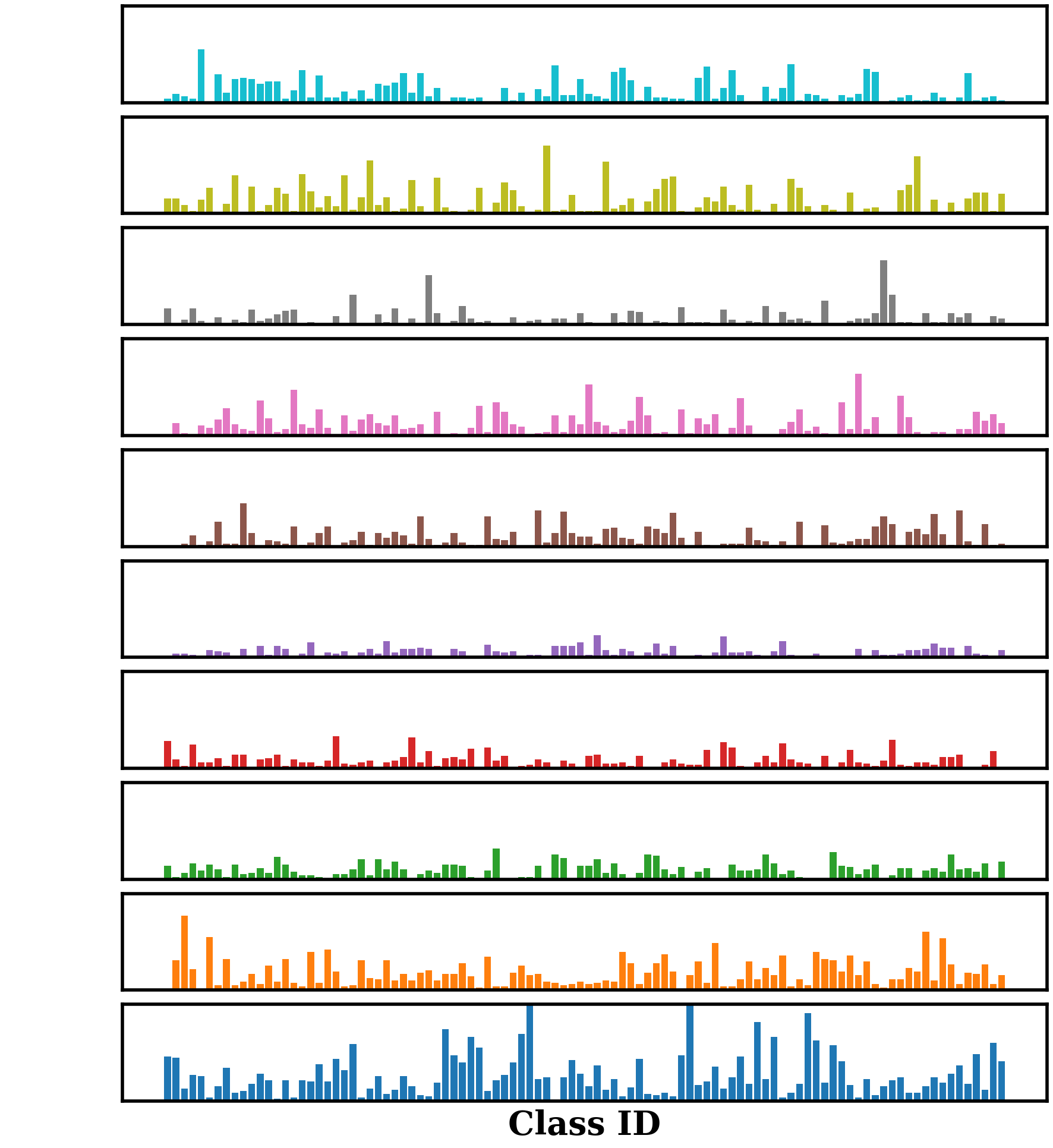}
         \caption{MVP~\citep{moon2023online}}
         \label{fig:mvp_selection_cnt}
     \end{subfigure}
     \hfill
     \begin{subfigure}{0.33\linewidth}
         \centering
         \includegraphics[width=\linewidth]{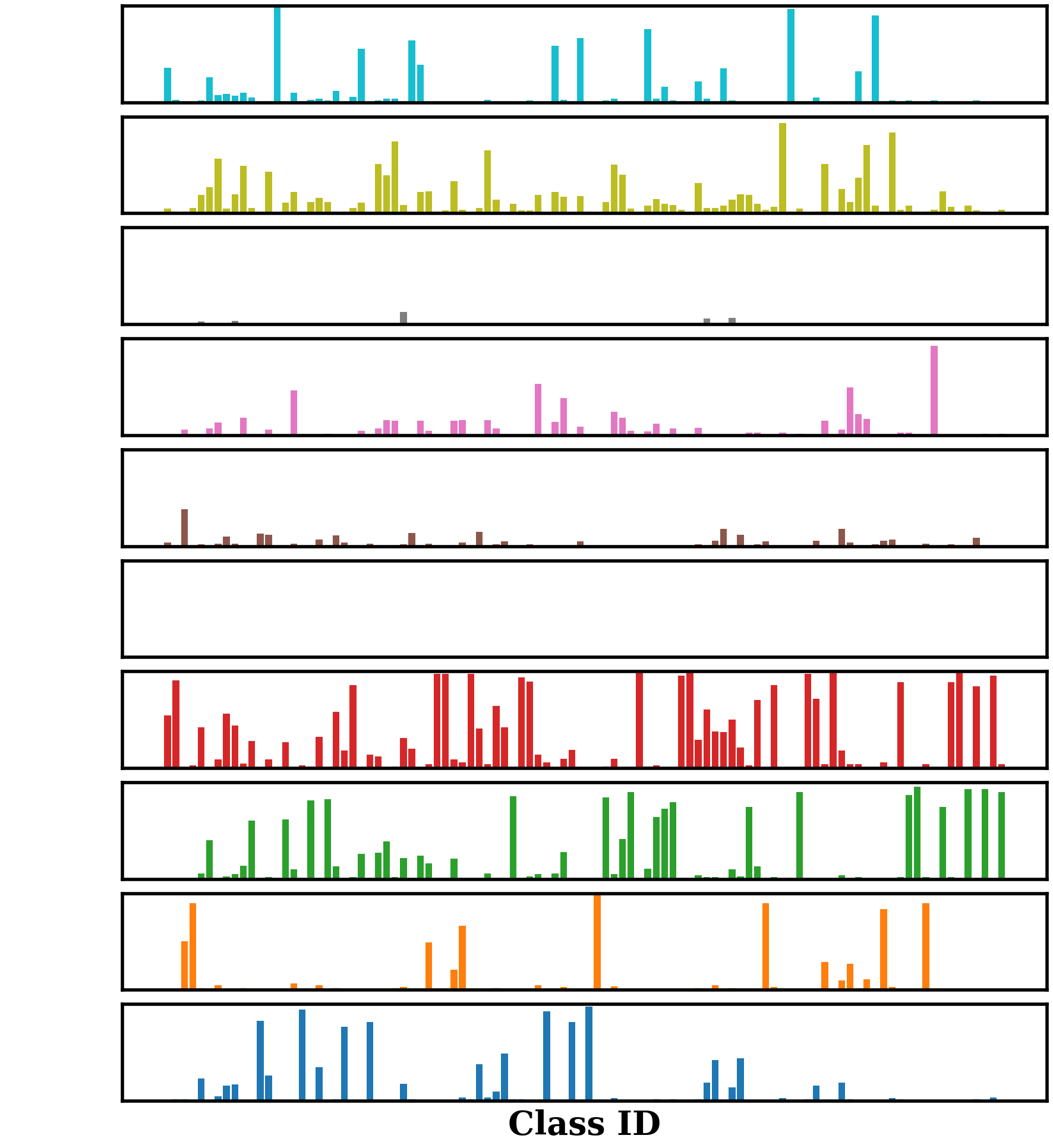}
         \caption{MISA~\citep{kang2025advancing}}
         \label{fig:misa_selection_cnt}
     \end{subfigure}
     \caption{Histograms of prompt selection counts per class on task-free continual learning using CIFAR100~\citep{krizhevsky2009learning}. In all histograms, the x-axis represents 100 class IDs and the y-axis denotes the number of times each prompt is selected. (a) L2P~\citep{wang2022learning} is evaluated on the 10-task class-incremental setting and the model selects the top-5 prompts per input from a prompt pool. (b) MVP~\citep{moon2023online} and (c) MISA~\citep{kang2025advancing} are evaluated on the 5-task Si-Blurry setting. Note that both methods select only one prompt per input image unlike L2P.}
     \label{fig:all-images}
\end{figure*}

\subsection{Parameter-Efficient Fine-Tuning for OCL} 

Recently, \textit{parameter-efficient fine-tuning (PEFT)} techniques have been actively explored in the context of online continual learning, as they leverage large-scale pretrained models with updating only a small number of learnable parameters. This enables models to adapt quickly to new data while significantly reducing computational costs. In practice, PEFT is typically implemented by introducing lightweight modules while keeping the backbone of the pretrained model frozen.

In computer vision, PEFT is often applied to a pretrained Vision Transformer (ViT)~\citep{dosovitskiy2021vit}, which consists of an input embedding layer $f_0:\mathcal{X}\rightarrow\mathbb{R}^{N\times D}$, and a stack of $L$ self-attention blocks $f_i:\mathbb{R}^{N\times D}\rightarrow\mathbb{R}^{N\times D},i=1,\ldots,L$ where $N$ is the number of tokens (including both patch and class tokens), $D$ and $L$ are the hidden dimension and the depth of the ViT, respectively. There are two common PEFT approaches for ViTs: (\textit{i}) low-rank adaptation~\citep{hu2022lora, wei2024online, zanella2024low}, which inserts learnable low-rank matrices into self-attention blocks, and (\textit{ii}) prompt tuning~\citep{lester-etal-2021-power}, which augments the input sequence with learnable tokens. A straightforward approach is to prepend $M$ learnable tokens to the encoder input sequence~\citep{wang2022learning}. Another variant, often referred to as prefix tuning, prepends learnable tokens only to the key and value sequences of certain self-attention blocks,
and it has also been widely adopted in the recent continual learning literature~\citep{wang2022dualprompt, roy_2024_CVPR, moon2023online, kang2025advancing}.
\section{Revisiting Prompt Selection in OCL}
\label{revisiting}

\subsection{What Is Prompt Selection?} 

As tasks continuously shift over time under OCL scenarios, using a single static prompt for all inputs may lead to suboptimal adaptation. Since different inputs and tasks may benefit from different prompts depending on their underlying distributions, dynamically selecting the most suitable prompt for each input enables more flexible and context-aware adaptation. This strategy, often referred to as \emph{prompt selection}, consists of two components: (\textit{i}) a \emph{prompt pool}, a set of learnable prompts, and (\textit{ii}) a \emph{selection mechanism}, which selects the optimal prompt based on the current input. The prompt pool is formally written as $\mathcal{P}=\{(\mathbf{p}_1,\mathbf{k}_1), \ldots, (\mathbf{p}_P,\mathbf{k}_P)\}$, where $\mathbf{p}_i\in\mathbb{R}^{M\times D}$ denotes a single prompt of $M$ learnable tokens, and $\mathbf{k}_i$ denotes a key vector associated with $\mathbf{p}_i$. Given an input $\mathbf{x}$, the selection mechanism first computes a query vector $\mathbf{q}$ from the input, then selects a prompt $\mathbf{p}_*$ from the pool $\mathcal{P}$ based on the similarities between the query $\mathbf{q}$ and the set of keys $\{\mathbf{k}_i\}$. This selection can be performed either through hard selection, where the prompt with the highest similarity is exclusively chosen, or through soft-routing, where multiple prompts are aggregated using their similarity scores. After selection, the prompt $\mathbf{p}_*$ is optimized during training and reused at inference time. 

An important remaining question is how to obtain query and key vectors to enable effective prompt selection. For queries, the most common approach is to simply use a pretrained encoder, such as the embedding vector of the class token from the final layer of a ViT~\citep{wang2022learning, wang2022dualprompt, moon2023online, kang2025advancing}. For keys, under task-based OCL scenarios, one can explicitly assign the $k$-th key $\mathbf{k}_k$ to the corresponding task $\mathcal{D}_k$. By maximizing the similarity between $\mathbf{k}_k$ and the query $\mathbf{q}$ obtained from an input $\mathbf{x}\in\mathcal{B}_t$, the same key is expected to be selected for inputs belonging to the task at inference time~\citep{wang2022dualprompt}. In contrast, under task-free scenarios, such explicit key-task assignment is infeasible since task IDs are unknown. As an alternative, the selection mechanism first chooses the key that is most similar to the query, and then updates the selected key to be better aligned with the current query~\citep{wang2022learning}. This approach does not rely on task IDs and is therefore well suited for the task-free continual learning.

\subsection{In-depth Analysis of Prompt Selection}
\label{sec:indepth_analysis}
In this section, we perform an in-depth analysis of representative prompt selection methods in continual learning. We focus on three task-free methods, L2P~\citep{wang2022learning}, MVP~\citep{moon2023online}, and MISA~\citep{kang2025advancing}, and two task-based methods, DualPrompt~\citep{wang2022dualprompt} and ConvPrompt~\citep{roy_2024_CVPR}. Note that MISA is the current state-of-the-art among task-free OCL methods.

Our goal is to examine whether the methods actually distinguish between inputs from different labels. We then revisit the necessity of prompt selection by evaluating its actual impact on performance.  
For the analysis, we use a prompt pool of $P=10$ prompts with CIFAR100~\citep{krizhevsky2009learning}, containing $C=100$ classes.\footnote{For ConvPrompt~\citep{roy_2024_CVPR}, the size of a prompt pool expands in a task-adaptive manner, reaching a final size of $P=50$.} Note that all the prompt selection methods are based on query-key similarity. The detailed experimental setup is described in the supplementary material.

\medskip
\noindent
\textbf{Analysis of Prior Prompt Selection Methods.}
We first examine task-free prompt selection methods. As described in the previous section, they increase the similarity between the input query and the selected key at each training iteration. After training, we evaluate the learned prompt pool by counting how often each prompt is selected for each class. Interestingly, we observe that the prompt selection appears to be largely uncorrelated with input features or class labels. Specifically, different prompts are often selected for samples from the same class. This leads to two issues: (\textit{i}) a single class may be associated with multiple prompts, causing its information to be fragmented, and (\textit{ii}) previously learned knowledge may be overwritten by data from later observed classes. For example, in L2P~\citep{wang2022learning}, we found no meaningful correlation between input labels and selected prompts, as shown in Figure~\ref{fig:l2p_selection_cnt}. No prompt is consistently associated with specific classes and the selection shows no class-specific bias across the 100 classes, indicating a lack of alignment between selected prompts and semantic labels. Although MVP~\citep{moon2023online} introduces contrastive prompt tuning to encourage separation among key vectors, it still exhibits similar uncorrelated behavior, as visualized in Figure~\ref{fig:mvp_selection_cnt}. MISA~\citep{kang2025advancing} shows a slightly better separation between prompts, but fails to fully utilize the prompt pool, with some prompts never or rarely selected (see Figure~\ref{fig:misa_selection_cnt}). These results suggest that existing prompt selection methods often fail to consistently associate prompts with specific input characteristics.\footnote{Similar failures occur in other datasets (see Supplementary Material).}

\begin{figure}[t]
     \centering
     \begin{subfigure}{0.49\linewidth}
         \centering
         \includegraphics[width=\linewidth]{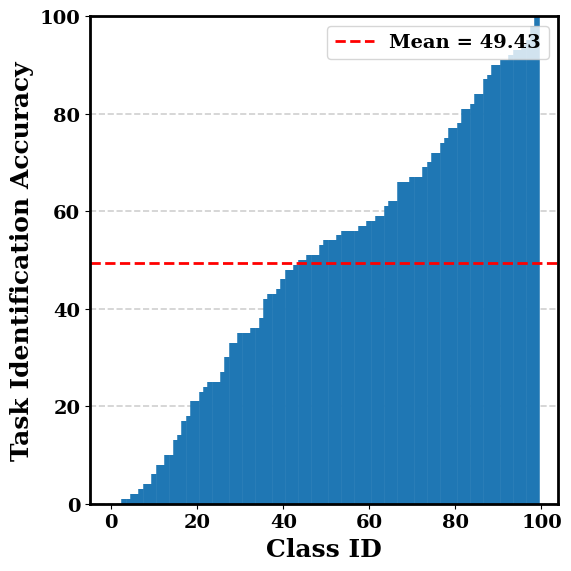}
         \caption{DualPrompt~\citep{wang2022dualprompt}}
         \label{fig:dual_acc}
     \end{subfigure}
     \hfill
     \begin{subfigure}{0.49\linewidth}
         \centering
         \includegraphics[width=\linewidth]{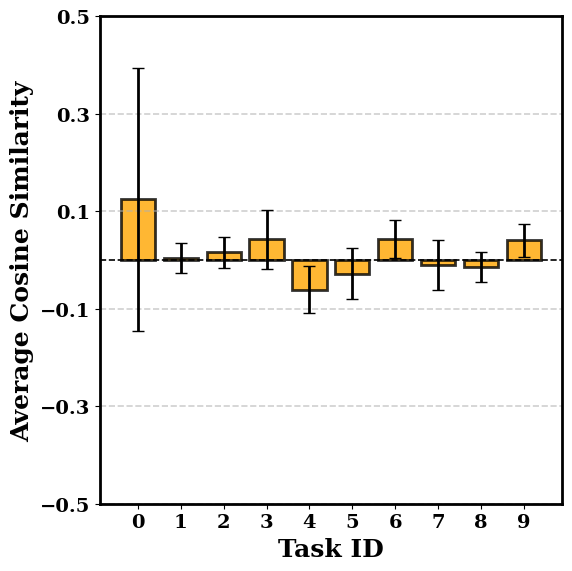}
         \caption{ConvPrompt~\citep{roy_2024_CVPR}}
         \label{fig:convprompt_cnt}
     \end{subfigure}
     \caption{Prompt selection failures in task-based methods on CIFAR100~\citep{krizhevsky2009learning}. (a) Task identification accuracy (\%) via prompt selection. The x-axis represents the 100 class IDs, sorted in ascending order of their task identification accuracy, which is shown on the y-axis as the proportion of correct prompt selection.
     (b) The average and standard variation of cosine similarities computed from the 7-th layer. The x-axis represents the task ID of input sample and the y-axis indicates the average cosine similarity between each task's samples and their assigned keys. Ideally, all value should be close to 1, but most are near 0. Note that consistent results are observed across other layers as well.}
     \label{fig:tbocl}
\end{figure}

Next, we also validate the effectiveness of the task-based method, DualPrompt~\citep{wang2022dualprompt} and ConvPrompt~\citep{roy_2024_CVPR}. In this setting, each prompt is explicitly assigned to a specific task using task IDs, allowing the model to focus training on the task-specific prompt. Consequently, the ability to correctly identify the prompt associated with the task of the input becomes crucial during evaluation. In DualPrompt, one prompt is assigned to each task, and only one prompt is selected based on the similarity score. Therefore, task identification accuracy (i.e., the proportion of correctly selecting prompt $i$ for task $i$ inputs) can be measured. As shown in Figure~\ref{fig:dual_acc}, the average accuracy was below 50\%, indicating that the model often failed to select the appropriate prompt at inference time.

In contrast, ConvPrompt assigns 5 prompts to each task and adopts a soft routing mechanism. To examine whether the prompts are appropriately selected, we measure the cosine similarity between the input and its assigned keys for each task. Ideally, it should be high across all tasks because the similarity directly contributes the selection of prompts.
However, as shown in Figure~\ref{fig:convprompt_cnt}, the similarities remain close to zero for all tasks, indicating that even with the soft routing approach, input-adaptive prompt selection does not function effectively.

\medskip
\noindent
\textbf{Revisiting Prompt Selection.}
These observations lead us to the following research question: \emph{Is prompt selection necessary for task-free online continual learning?} To answer this, we conduct ablation experiments on the Si-Blurry task-free OCL setup~\citep{moon2023online} using CIFAR100 dataset~\citep{krizhevsky2009learning}, with five random seeds to ensure robustness. We use the prompt selection method described in the previous section, which selects a prompt based on query-key similarity and updates the selected key to be better aligned with the input query. Following recent frameworks~\citep{wang2022dualprompt,kang2025advancing}, we use a single shared prompt per layer in the first two self-attention layers, and use a prompt pool of $10$ candidates per layer in the next three layers.

\begin{table}[t]
\caption{Ablation study on the prompt selection. Note that (a) is same as MISA~\citep{kang2025advancing} without pretrained prompts.}
\label{tab:prompt_selection_fail}
\centering
\setlength{\tabcolsep}{1mm}
\resizebox{\linewidth}{!}{%
\begin{tabular}{cccccc}
\toprule
& {\#\ of Prompts} & {Selection} & {Params} & $A_\text{auc}$ (↑) & $A_\text{last}$ (↑) \\
\midrule
(a) & 10 & Similarity-based & 560K & 79.89\small{$\pm$2.14} & 82.12\small{$\pm$1.66} \\ 
(b) & 10 & Random & 553K & 79.78\small{$\pm$2.03} & 81.85\small{$\pm$1.78} \\ 
(c) & 1 & N/A & 138K & \textbf{81.05}\small{$\pm$2.18} & \textbf{82.38}\small{$\pm$1.95} \\ 
\bottomrule
\end{tabular}}
\end{table}

We first examine the key-based selection mechanism. To this end, we replace the similarity-based selection scheme with random selection. Interestingly, as shown in the first two rows (a-b) in Table~\ref{tab:prompt_selection_fail}, random selection leads to only a negligible performance drop compared to the learned selection method, suggesting that key-based prompt selection may not be as beneficial as expected.

Given the limited benefit of key-based selection, we further examine the necessity of using multiple prompts. To this end, we simplify the architecture by using only one prompt per layer. As shown in the last row (c) of Table~\ref{tab:prompt_selection_fail}, this configuration improves overall accuracy while reducing the number of learnable parameters by approximately 75\%. These findings suggest that a single prompt can serve as a strong alternative for task-free online continual learning. 
\section{Proposed Method: SinglePrompt}
\label{method}

\begin{table*}[t]
\caption{Accuracy (\%) on CIFAR100~\citep{krizhevsky2009learning}, Tiny-Imagenet~\citep{le2015tiny}, Imagenet-R~\citep{hendrycks2021many}. We average the performance over 5 random seeds ($\text{Mean} \pm \text{Std}$) for ours. Others reported numbers borrow from MISA~\citep{kang2025advancing}. \textbf{Bold} indicates the best performance, while underlined entries denote the second-best results.}
\label{tab:main_table}
\centering

\scalebox{0.85}{%
\begin{tabular}{llcccccc}
\toprule
 & & \multicolumn{2}{c}{CIFAR100}
& \multicolumn{2}{c}{Tiny-ImageNet}
& \multicolumn{2}{c}{ImageNet-R} \\
\cmidrule(lr){3-4}
\cmidrule(lr){5-6}
\cmidrule(lr){7-8} 
Buffer & Method & $A_\text{auc}$ (↑) & $A_\text{last}$ (↑) & $A_\text{auc}$ (↑) & $A_\text{last}$ (↑) & $A_\text{auc}$ (↑) & $A_\text{last}$ (↑) \\
\midrule 
\multirow{9}{*}{0} & Finetuning  & 19.71\small{$\pm$3.39} & 10.42 \small{$\pm$4.92} & 15.50 \small{$\pm$0.74} & 10.42\small{$\pm$4.92} & 7.51\small{$\pm$3.94} & 2.29 \small{$\pm$0.85}  \\
    &Linear Probe    & 49.69\small{$\pm$6.09} & 23.07\small{$\pm$7.33} & 42.15\small{$\pm$2.79} & 21.97\small{$\pm$6.43} & 29.24\small{$\pm$1.26} & 16.87\small{$\pm$3.14}  \\
    &EWC~\citep{kirkpatrick2017overcoming}
    & 49.51\small{$\pm$0.52}  & 52.83\small{$\pm$2.31}  & 51.70\small{$\pm$2.89} & 31.04\small{$\pm$3.12} & 31.58\small{$\pm$1.04} & 20.72\small{$\pm$1.11}  \\
    &LwF~\citep{li2017learning}       
    & 55.51\small{$\pm$3.49} & 36.53\small{$\pm$10.96} & 49.00\small{$\pm$1.52} & 27.47\small{$\pm$7.59} & 31.61\small{$\pm$1.53} & 20.62\small{$\pm$3.67} \\
    &L2P~\citep{wang2022learning}        
    & 57.08\small{$\pm$4.43} & 41.63\small{$\pm$12.73} & 52.09\small{$\pm$1.92} & 35.05\small $\pm${5.73} & 29.65\small{$\pm$1.63} & 19.55\small{$\pm$4.78}  \\
    &DualPrompt~\citep{wang2022dualprompt}     
    & 67.07\small{$\pm$4.16} & 56.82\small{$\pm$3.49} & 66.09\small{$\pm$2.00} & 48.72\small{$\pm$3.41} & 40.11\small{$\pm$1.27} & 29.24\small{$\pm$4.63}  \\
    &MVP~\citep{moon2023online}
    &  68.10\small{$\pm$4.91} & 62.59\small{$\pm$2.38} & 68.95\small{$\pm$1.33} & 52.78\small $\pm${2.08} & 40.60\small{$\pm$1.21} & 31.96\small{$\pm$3.07}  \\
    &MISA~\citep{kang2025advancing}
    & \underline {80.55}\small{$\pm$2.17} & \underline {80.98}\small{$\pm$1.08} & \underline {80.44}\small{$\pm$0.80} & \underline {74.84}\small{$\pm$0.64} & \underline {50.89}\small{$\pm$1.03} & \underline {43.92}\small{$\pm$0.37} \\
    & \cellcolor{gray!15}\textbf{SinglePrompt} (ours) & \cellcolor{gray!15}\textbf{85.58}\small{$\pm$1.46} & \cellcolor{gray!15}\textbf{87.53}\small{$\pm$0.45} & \cellcolor{gray!15}\textbf{85.60}\small{$\pm$0.59} & \cellcolor{gray!15}\textbf{82.87}\small{$\pm$0.54} & \cellcolor{gray!15}\textbf{59.00}\small{$\pm$1.16} & \cellcolor{gray!15}\textbf{58.88}\small{$\pm$0.53} \\
\midrule
    \multirow{8}{*}{500} & ER~\citep{rolnick2019experience}      
    & 65.57\small{$\pm$4.77} & 60.68\small{$\pm$1.15} & 59.46\small{$\pm$1.81} & 40.60\small{$\pm$2.71} & 40.31\small{$\pm$1.33} & 28.85\small{$\pm$1.43}\\
    &DER++~\citep{buzzega2020dark}       
    & 66.92\small{$\pm$4.16}  & 65.63\small{$\pm$0.72} & 61.67\small{$\pm$1.19} & 46.03\small{$\pm$1.00} &40.32\small{$\pm$1.08} & 31.53\small{$\pm$1.57}  \\
    &ER-ACE~\citep{caccia2021new}       
    & 69.36\small{$\pm$3.01}  & 72.07\small{$\pm$0.62} & 64.52\small{$\pm$0.78} & 56.82\small{$\pm$0.67} & 41.06\small{$\pm$1.32} & 36.59\small{$\pm$0.52}  \\
    &RM~\citep{Bang_2021_CVPR}      
    & 40.86\small{$\pm$3.32} & 23.94\small{$\pm$0.61} & 31.96\small{$\pm$0.80} & 7.43\small{$\pm$0.27} & 18.31\small{$\pm$1.09} & 4.14\small{$\pm$0.18}\\
    &CLIB~\citep{koh2022online}      
    & 69.68\small{$\pm$2.20} & 67.16\small{$\pm$0.72} & 60.11\small{$\pm$1.53} & 48.97\small{$\pm$1.48} & 37.18\small{$\pm$1.52} & 29.51\small{$\pm$0.98}\\
    &MVP~\citep{moon2023online}     
    & 76.06\small{$\pm$4.22} & 79.32\small{$\pm$1.28} & 76.52\small{$\pm$0.73} & 65.19\small{$\pm$0.58} & 49.07\small{$\pm$1.47} & 44.17\small{$\pm$1.72}\\
    &MISA~\citep{kang2025advancing}
    & \underline {82.37}\small{$\pm$1.54} & \underline {82.27}\small{$\pm$ 0.73} & \underline {79.08}\small{$\pm$ 0.60}& \underline {69.91}\small{$\pm$ 0.52} & \underline {54.72}\small{$\pm$1.15} & \underline {47.48}\small{$\pm$0.57} \\
    & \cellcolor{gray!15}\textbf{SinglePrompt} (ours) & \cellcolor{gray!15}\textbf{84.34}\small{$\pm$1.21} & \cellcolor{gray!15}\textbf{85.61}\small{$\pm$0.55} &  \cellcolor{gray!15}\textbf{82.69}\small{$\pm$0.28} & \cellcolor{gray!15}\textbf{74.45}\small{$\pm$0.72} & \cellcolor{gray!15}\textbf{58.35}\small{$\pm$1.57} & \cellcolor{gray!15}\textbf{53.45}\small{$\pm$0.90} \\
\midrule
    \multirow{8}{*}{2000} & ER~\citep{rolnick2019experience}        
    & 69.86\small{$\pm$4.08} & 71.81\small{$\pm$0.69} & 66.75\small{$\pm$1.13} & 55.07\small{$\pm$1.28} & 45.74\small{$\pm$1.35} & 38.13\small{$\pm$0.32}\\
    &DER++~\citep{buzzega2020dark}       
    & 69.42\small{$\pm$3.65}  & 65.68\small{$\pm$0.72} & 66.58\small{$\pm$0.88} & 56.81\small{$\pm$0.65} &42.79\small{$\pm$1.31} & 36.06\small{$\pm$1.04}  \\
    &ER-ACE~\citep{caccia2021new}       
    & 70.59\small{$\pm$3.02}  & 74.75\small{$\pm$0.19} & 66.86\small{$\pm$0.84} & 58.40\small{$\pm$0.38} & 43.62\small{$\pm$1.31} & 40.49\small{$\pm$0.22}  \\
    &RM~\citep{Bang_2021_CVPR}      
    & 53.27\small{$\pm$3.00} & 65.51\small{$\pm$0.55} & 47.26\small{$\pm$1.13} & 44.55\small{$\pm$0.37} & 27.88\small{$\pm$1.29} & 24.25\small{$\pm$0.99} \\
    &CLIB~\citep{koh2022online}     
    & 71.53\small{$\pm$2.61} & 72.09\small{$\pm$0.49} & 65.47\small{$\pm$0.76} & 56.87\small{$\pm$0.54} & 42.69\small{$\pm$1.30} & 35.43\small{$\pm$0.38}\\
    &MVP~\citep{moon2023online}     
    & 78.65\small{$\pm$3.59} & 84.42\small{$\pm$0.44} & 80.67\small{$\pm$0.75} & 74.34\small{$\pm$0.32} & 52.47\small{$\pm$1.45} & 50.54\small{$\pm$2.08}\\
    &MISA~\citep{kang2025advancing}
    & \underline {83.58}\small{$\pm$1.72} & \underline {85.32}\small{$\pm$0.25} & \underline {82.91}\small{$\pm$0.47} & \underline{76.41}\small{$\pm$0.33} & \underline {57.67}\small{$\pm$0.72} & \underline {53.62}\small{$\pm$0.68} \\
    & \cellcolor{gray!15}\textbf{SinglePrompt} (ours) & \cellcolor{gray!15}\textbf{86.07}\small{$\pm$1.76} & \cellcolor{gray!15}\textbf{88.34}\small{$\pm$0.31} &  \cellcolor{gray!15}\textbf{85.69}\small{$\pm$0.43} & \cellcolor{gray!15}\textbf{81.67}\small{$\pm$0.20} & \cellcolor{gray!15}\textbf{61.60}\small{$\pm$1.36} & \cellcolor{gray!15}\textbf{59.87}\small{$\pm$0.46} \\
\bottomrule
\end{tabular}}
\end{table*}

Based on our findings introduced in the previous section, we now propose \textbf{SinglePrompt}, a simple yet effective framework that eliminates prompt selection for task-free online continual learning. Specifically, since our observations reveal that the task-adaptive prompt pool methods do not function effectively, we focus on improving performance on the classifier side rather than on the prompts. Our framework consists of three simple components: \emph{(i) prefix tuning for online adaptation}, \emph{(ii) logit calculation via cosine similarity}, and \emph{(iii) classification with minibatch logit masking}. See Figure~\ref{fig:main_fig} for an overview of our framework.

\medskip
\noindent
\textbf{Prefix Tuning for Online Adaptation.}
We apply prefix-tuning~\citep{li_2021_prefix} to the first $K$ self-attention blocks of a pretrained ViT. For each block $f_i,i\in\{1,\ldots,K\}$, learnable prompts $\mathbf{p}_i^k,~\mathbf{p}_i^v \in \mathbb{R}^{M \times D}$ are prepended to the key and value sequences, respectively. Specifically, the attention operation in the block $f_i$ is modified from
\begin{equation}
    \text{Attn}(\mathbf{Q}, \mathbf{K}, \mathbf{V})=\text{softmax}\Big( \frac{\mathbf{Q}\mathbf{K}^\top}{\sqrt{D}} \Big)\mathbf{V}
    \label{eq:attn_ori}
\end{equation}
to
\begin{equation}
    \text{Attn}(\mathbf{Q},\mathbf{K},\mathbf{V};\mathbf{p}_i^k,\mathbf{p}_i^v)= \\
    \text{softmax}\Big( \frac{\mathbf{Q}[\mathbf{p}_i^k;\mathbf{K}]^\top}{\sqrt{D}} \Big)[\mathbf{p}_i^v;\mathbf{V}],
    \label{eq:attn_prefix}
\end{equation}
where $\mathbf{Q},\mathbf{K}$, and $\mathbf{V}$ are query, key, and value matrices, respectively. After passing through all $L$ blocks, the class token is used for the final representation $\mathbf{g}\in\mathbb{R}^D$.

\begin{figure}[t]
    \centering
    \includegraphics[width=0.9\linewidth]{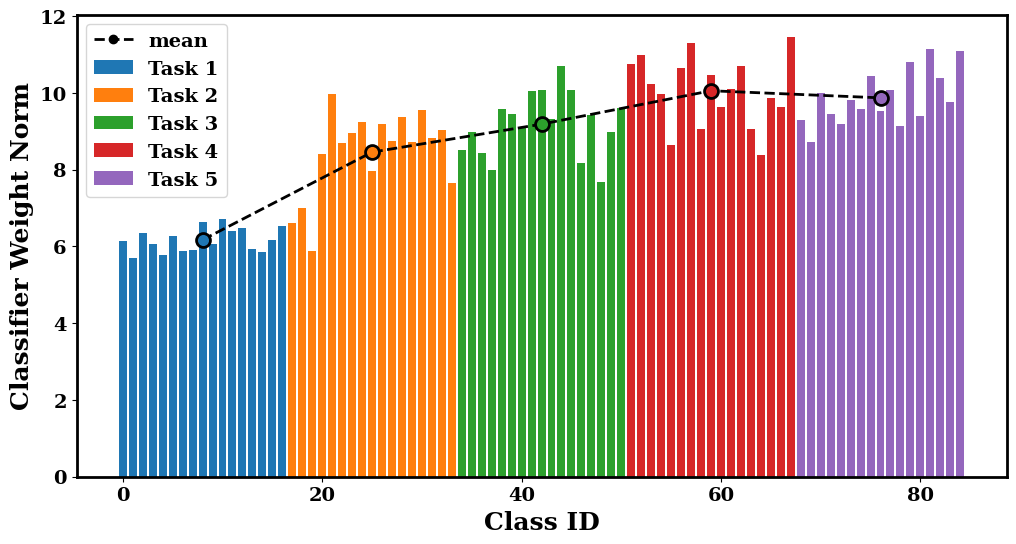}
    \vspace{-0.1in}
    \caption{Visualization of the L2 norms of the weights for each class of the linear classifier, with blurry classes excluded. The x-axis shows class IDs in the order the model observes them.}
    \label{fig:head_weight}
\end{figure}

\medskip
\noindent
\textbf{Logit Calculation via Cosine Similarity.}
For the linear classifier commonly used in image classification, we observed that the weights of classes learned in the early stages of continual learning tend to have relatively smaller L2 norms (see Figure~\ref{fig:head_weight}). Since weight norm directly affects the final prediction score, this imbalance can lead to forgetting. To address this issue, we adopt a cosine similarity-based logit computation, where the norm has no influence in the final prediction score. We denote the prototype of class $c$ as $\mathbf{c}_c\in\mathbb{R}^D$. Given a feature vector $\mathbf{g}\in\mathbb{R}^D$, the cosine similarity between $\mathbf{g}$ and $\mathbf{c}_c$ is computed as: 
\begin{equation}
    \mathbf{z}_c=\frac{\mathbf{g} \cdot \mathbf{c}_c}{\|\mathbf{g}\| \|\mathbf{c}_c\|} \cdot \frac{1}{\tau},
    \label{eq:final_logit}
\end{equation}
where $\tau$ denotes a temperature hyperparameter, set to $0.1$ in our experiments.
The vector $\mathbf{z}\in\mathbb{R}^C$ serves as the final logits at inference time.

\medskip
\noindent
\textbf{Classification with Minibatch Logit Masking.}
Following MISA~\citep{kang2025advancing}, we employ minibatch-wise logit masking, which zeros out the probabilities of classes not presented in the current minibatch $\mathcal{B}_t=\{(\mathbf{x}_t^{(i)}, y_t^{(i)})\}_{i=1}^{B_t}$ by adding a mask vector $\mathbf{m}\in\mathbb{R}^C$ to the logits, where
\begin{equation}
  \mathbf{m}_c=\begin{cases} 0 & \text{if }c\in\{y_t^{(i)}|i=1, \cdots,B_t\}, \\
  -\infty & \text{otherwise}.\end{cases}
  \label{eq:logit_masking}
\end{equation}
The masked logit $\tilde{\mathbf{z}}^{(i)}=\mathbf{z}^{(i)}+\mathbf{m}^{(i)} \in \mathbb{R}^C$ for each sample $\mathbf{x}^{(i)}$ from $\mathcal{B}_t$ is then used to compute the cross-entropy loss $\mathcal{L}_\mathtt{CE}$ for the current minibatch $\mathcal{B}_t$:
\begin{equation}
    \mathcal{L}_\mathtt{CE}=-\frac{1}{B_t}\sum_{i=1}^{B_t}\sum_{c=1}^C  \mathbf{1}_{[y_t^{(i)}=c]}\log\frac{\exp(\tilde{\mathbf{z}}^{(i)}_c)}{\sum_{c'} \exp(\tilde{\mathbf{z}}^{(i)}_{c'})} ,
\end{equation}
where $\mathbf{1}_{[y_t^{(i)}=c]}$ denotes the indicator function. Note that this logit masking technique prevents the class prototypes of absent classes from being updated during training. At inference time, however, logit masking is not applied because we have no access to the ground-truth labels of the current minibatch.

\section{Experiments}
\label{experiments}

\begin{table*}[t]
\caption{Ablation studies on different PEFT Adapters, logit types, batch masking and K values conducted on CIFAR100~\citep{krizhevsky2009learning}. The prompt length is fixed at 20 in all experiments involving prompts. We average the performance over 5 random seeds ($\text{Mean} \pm \text{Std}$). \textbf{Bold} indicates the best result within each ablation setting.}
\label{table:combined_experiments}
\centering
\scalebox{0.85}{
\begin{tabular}{ccccc|c|c|c}
\toprule
PEFT Adapter & Logit type & Batch Masking & K & $\#\text{Params}$ & $A_\text{auc}$ (↑) & $A_\text{last}$ (↑) & $F_\text{last}$ (↓)\\
\hline
\cellcolor{red!15}Prompt tuning & cosine similarity & \checkmark & N/A & 92K & 81.68\small{$\pm$1.34} & 84.37\small{$\pm$0.38} & \textbf{5.68}\small{$\pm$0.92}\\
\cellcolor{red!15}LoRA & cosine similarity & \checkmark & N/A & 138K & 82.21\small{$\pm$1.74} & 81.25\small{$\pm$1.64} & 11.45\small{$\pm$0.76} \\
\cellcolor{red!15}AdaptMLP & cosine similarity & \checkmark & N/A & 203K & 83.74\small{$\pm$1.97} & 84.34\small{$\pm$0.42} & 9.07\small{$\pm$0.66} \\
\rowcolor{gray!15}
\cellcolor{red!15}Prefix tuning & cosine similarity & \checkmark & 5 & 230K & \textbf{85.58}\small{$\pm$1.46} & \textbf{87.53}\small{$\pm$0.45} & 5.89\small{$\pm$0.99}  \\
\hline
Prefix tuning & \cellcolor{blue!15}linear classifier & \checkmark & 5 & 230K & 81.63\small{$\pm$1.98} & 82.37\small{$\pm$1.75} & 10.82\small{$\pm$1.57} \\
\rowcolor{gray!15}

Prefix tuning & \cellcolor{blue!15}cosine similarity & \checkmark & 5 & 230K & \textbf{85.58}\small{$\pm$1.46} & \textbf{87.53}\small{$\pm$0.45} & \textbf{5.89}\small{$\pm$0.99} \\
\hline
Prefix tuning & cosine similarity & \cellcolor{green!15}\ding{55} & 5 & 230K & 65.57\small{$\pm$5.17} & 51.81\small{$\pm$3.39} & 48.19\small{$\pm$3.89} \\
\rowcolor{gray!15}
Prefix tuning & cosine similarity & \cellcolor{green!15}\checkmark & 5 & 230K & \textbf{85.58}\small{$\pm$1.46} & \textbf{87.53}\small{$\pm$0.45} & \textbf{5.89}\small{$\pm$0.99} \\
\hline
Prefix tuning & cosine similarity & \checkmark & \cellcolor{orange!15}3 & 168K & 85.09\small{$\pm$1.51} & 87.04\small{$\pm$0.24} & 6.07\small{$\pm$0.67} \\
Prefix tuning & cosine similarity & \checkmark & \cellcolor{orange!15}4 & 199K & 85.13\small{$\pm$1.57} & 87.37\small{$\pm$0.38} & 5.90\small{$\pm$1.01} \\
\rowcolor{gray!15}
Prefix tuning & cosine similarity & \checkmark & \cellcolor{orange!15}5 & 230K & 85.58\small{$\pm$1.46} & \textbf{87.53}\small{$\pm$0.45} & \textbf{5.89}\small{$\pm$0.99} \\
Prefix tuning & cosine similarity & \checkmark & \cellcolor{orange!15}6 & 261K & \textbf{85.72}\small{$\pm$1.50} & 87.39\small{$\pm$0.55} & 6.16\small{$\pm$0.71} \\
Prefix tuning & cosine similarity & \checkmark & \cellcolor{orange!15}9 & 291K & 84.93\small{$\pm$1.79} & 85.76\small{$\pm$0.84} & 8.51\small{$\pm$1.78} \\
Prefix tuning & cosine similarity & \checkmark & \cellcolor{orange!15}12 & 322K & 84.97\small{$\pm$1.80} & 86.09\small{$\pm$0.21} & 8.33\small{$\pm$0.98} \\
\bottomrule
\end{tabular}}
\end{table*}

\subsection{Experimental Details}
\textbf{Baselines.} 
 We compare our approach against a broad set of continual learning baselines, including replay-based methods: ER~\citep{rolnick2019experience}, DER++~\citep{buzzega2020dark}, ER-ACE~\citep{caccia2021new}, Rainbow Memory(RM)~\citep{Bang_2021_CVPR}, CLIB~\citep{koh2022online}, regularization-based methods: LwF~\citep{li2017learning}, EWC~\citep{kirkpatrick2017overcoming}, and prompt-based methods: L2P~\citep{wang2022learning}, DualPrompt~\citep{wang2022dualprompt}. We also include comparisons with Si-Blurry-specific methods such as MVP~\citep{moon2023online} and MISA~\citep{kang2025advancing}. All baseline results are from MISA~\citep{kang2025advancing}.

\medskip
\noindent
\textbf{Setup.} 
We evaluate our method on three representative datasets: CIFAR100~\citep{krizhevsky2009learning}, Tiny ImageNet~\citep{le2015tiny}, and ImageNet-R~\citep{hendrycks2021many}, consisting of approximately 60K, 100K, and 30K samples and 100, 200, and 200 classes, respectively. 
To ensure fairness, we follow the implementation details of prior methods~\citep{moon2023online, kang2025advancing}. Specifically, we set the disjoint class ratio to 50\% and the blurry sample ratio to 10\% and use the Adam optimizer with a learning rate of 0.005 and a batch size of 32. The memory buffer is also configured in the same manner as in previous works. All experiments are conducted using the same ImageNet-pretrained ViT-B/16 backbone with a unified input resolution of 224. Unless otherwise specified, results are averaged over five runs with different random seeds. 

\medskip
\noindent
\textbf{Evaluation Metrics.}
Following prior works in task-free online continual learning~\citep{moon2023online, wei2024online, kang2025advancing}, we use three metrics for evaluation, $A_{\text{auc}}$, $A_{\text{last}}$ and $F_{\text{last}}$. $A_{\text{auc}}$~\citep{koh2022online} evaluates the performance of any-time inference by computing the area under the accuracy curve throughout the entire training stream, making it well suited for the task-free and online learning setup. $A_{\text{last}}$ denotes the accuracy at the end of the training, providing the final model performance. $F_{\text{last}}$ measures the forgetting after training all tasks, which is the performance degradation on previously learned classes. The precise definition and computation details of these metrics are provided in the supplementary material.

\subsection{Main Results}
We compare our proposed simple baseline to prior works in Online Continual Learning. The results are shown in Table~\ref{tab:main_table}. Our approach consistently achieves the best performance across all datasets. On CIFAR100~\citep{krizhevsky2009learning}, our method outperforms the prior state-of-the-art MISA~\citep{kang2025advancing}, with approximately 60\% fewer learnable parameters, by 6.55\% in $A_\text{last}$. Furthermore, our approach yields consistent gains of 8.03\% and 14.96\% in $A_\text{last}$ on Tiny-ImageNet~\citep{le2015tiny} and ImageNet-R~\citep{hendrycks2021many}, respectively, confirming its robustness across diverse datasets. 
Our method also achieves the highest performance by a considerable margin in all experiments conducted with the memory buffer.

Furthermore, we conduct online continual learning experiments by varying the disjoint class ratio not only to 50\% but also to 0\%, 25\%, 75\%, and 100\%, covering scenarios from the single-task setting (i.e., the disjoint class ratio is 0\%) to the fully disjoint setting (i.e., the disjoint class ratio is 100\%). The results are reported in Table~\ref{tab:disjointclassratio}. Compared to the prior art, MISA~\citep{kang2025advancing}, our method consistently achieves higher performance across all the scenarios, which demonstrates our robustness depending on the scenarios.

\begin{table}[t]
\caption{Comparison study on disjoint class ratios conducted on CIFAR100~\citep{krizhevsky2009learning}. We average the performance over 5 random seeds ($\text{Mean} \pm \text{Std}$). \textbf{Bold} indicates the best result.}
\label{tab:disjointclassratio}
\centering
\resizebox{\linewidth}{!}{%
\begin{tabular}{cccc}
\toprule
Disjoint class ratio (\%) & Method & $A_\text{auc}$ (↑) & $A_\text{last}$ (↑) \\
\midrule
\multirow{2}{*}{0} & MISA~\citep{kang2025advancing} & 77.91\small{$\pm$1.63} & 82.26\small{$\pm$0.92} \\
& \cellcolor{gray!15}SinglePrompt & \cellcolor{gray!15}\textbf{83.26}\small{$\pm$0.89} & \cellcolor{gray!15}\textbf{88.15}\small{$\pm$0.18} \\
\hline
\multirow{2}{*}{25} & MISA~\citep{kang2025advancing} & 79.13\small{$\pm$2.55} & 81.36\small{$\pm$1.01} \\
& \cellcolor{gray!15}SinglePrompt & \cellcolor{gray!15}\textbf{84.65}\small{$\pm$2.29} & \cellcolor{gray!15}\textbf{87.67}\small{$\pm$0.52} \\
\hline
\multirow{2}{*}{50} & MISA~\citep{kang2025advancing} & 80.55\small{$\pm$2.17} & 80.98\small{$\pm$1.08} \\
& \cellcolor{gray!15}SinglePrompt & \cellcolor{gray!15}\textbf{85.58}\small{$\pm$1.46} & \cellcolor{gray!15}\textbf{87.53}\small{$\pm$0.45} \\
\hline
\multirow{2}{*}{75} & MISA~\citep{kang2025advancing} & 82.16 \small{$\pm$1.79} & 80.98\small{$\pm$1.02} \\
& \cellcolor{gray!15}SinglePrompt & \cellcolor{gray!15}\textbf{87.16}\small{$\pm$0.94} & \cellcolor{gray!15}\textbf{86.80}\small{$\pm$0.56} \\
\hline
\multirow{2}{*}{100} & MISA~\citep{kang2025advancing} & 85.02 \small{$\pm$1.43} & 80.46\small{$\pm$1.28} \\
& \cellcolor{gray!15}SinglePrompt & \cellcolor{gray!15}\textbf{89.63}\small{$\pm$1.36} & \cellcolor{gray!15}\textbf{86.89}\small{$\pm$0.92} \\
\bottomrule
\end{tabular}}
\end{table}

\medskip
\noindent
\subsection{Ablation Studies}
\colorbox{red!15}{\textbf{PEFT Adatper.}}
Although our primary focus is on the prefix tuning strategy, we additionally conduct experiments with other parameter-efficient fine-tuning approaches, including LoRA, AdaptMLP, and prompt tuning for a more comprehensive comparison. All experiments are conducted on CIFAR100~\citep{krizhevsky2009learning}. For the LoRA experiment, we adopt the model architecture proposed in prior work on Online-LoRA~\citep{wei2024online}. Specifically, LoRA modules are applied to the query and value projection matrices, using a single shared LoRA instead of incrementally increasing parameters as in the original study. For AdaptMLP~\citep{chen2022adaptformer}, we integrate a 2-layer MLP in parallel with the original MLP block within each self-attention layer. In the case of prompt tuning, we employ a single prompt that is propagated from the input layer through the final output. The prompt length used in the prompt tuning experiment is kept identical as 20 to ensure a fair comparison with the prefix tuning. As shown in Table \ref{table:combined_experiments}, among the single adapter approaches, the prefix tuning method achieves the best performance.

\medskip
\noindent
\colorbox{blue!15}{\textbf{Logit Type.}}
Next, we validate the effectiveness of using cosine similarity logits. Compared to the experiment that employs a linear logit type (i.e., $\textbf{g}W^\top+b$ where $W \in \mathbb{R}^{C \times D}$ is a weight matrix and $b \in \mathbb{R}^C $ is a bias), our method achieves consistent improvement across all metrics, specifically improving $F_\text{last}$ by 4.93\%. This result suggests that the imbalance in weight norm,  as shown in Figure~\ref{fig:head_weight}, induces forgetting and consequently results in a performance drop.

\medskip
\noindent
\colorbox{green!15}{\textbf{Batch Masking.}}
We observe that removing minibatch logit masking also leads to a significant degradation in performance, indicating that allowing model to access only the class prototypes corresponding to the classes present in current minibatch is beneficial.

\begin{figure}[t]
    \centering
    \includegraphics[width=0.8\linewidth]{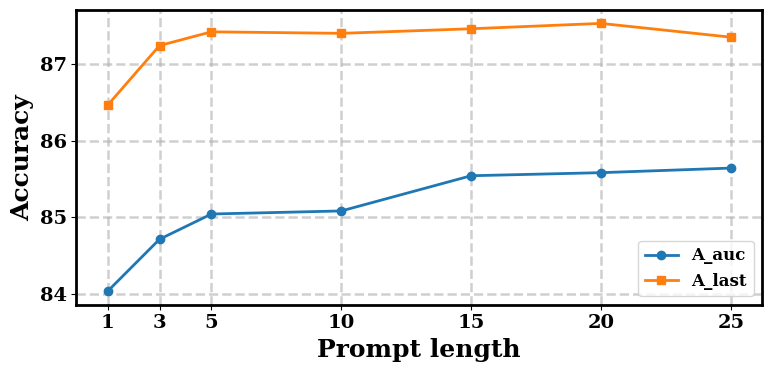}
    \caption{Ablation study on Prompt length. The x-axis denotes the value of $M$.}
    \label{fig:promptlengthAblation}
\end{figure}

\medskip
\noindent
\colorbox{orange!15}{\textbf{Prompt Injected Layers.}}
We examine the effect of the number of layers where prefix tuning is applied in Table~\ref{table:combined_experiments}. Specifically, we evaluate performance by prefix tuning at the $i$-th layer where $i=1,\ldots,K$. The performance improves as $K$ increases up to 5, but drops for $K=9$ and $K=12$. Furthermore, although $K=6$ yields results comparable to $K=5$, it requires more learnable parameters, which motivates the final choice of $K=5$.

\medskip
\noindent
\textbf{Prompt Length.}
Lastly, we investigate the effect of prompt length $M$. As shown in Figure~\ref{fig:promptlengthAblation}, Both $A_\text{auc}$ and $A_\text{last}$ improve as $M$ increases up to 20. However when $M>20$, $A_\text{auc}$ reaches saturation and $A_\text{last}$ exhibits a slight degradation in performance. Therefore, we set $M=20$ in our experiments.

\section{Related Works}
\label{related_works}

\textbf{Prompt-Based Methods.}
In conventional continual learning, prompt-based methods can be broadly categorized into two paradigms: task-specific training~\citep{wang2022dualprompt, roy_2024_CVPR} and prompt pool utilization~\citep{wang2022learning}. These approaches differ primarily in how prompts are selected. Task-specific training relies on known task IDs, enabling the model to associate each task with a dedicated prompt. In contrast, prompt pool methods operate in a task-agnostic manner by learning instance-wise prompt queries. Each prompt in the pool is associated with a key, and prompts are selected based on the similarity between the input query and the prompt keys. This mechanism makes prompt pool methods particularly suitable for \emph{task free OCL} scenarios, where task boundaries or IDs are not provided.

\medskip
\noindent
\textbf{Task-Free Online Continual Learning.}
Task-free online continual learning has emerged as a realistic setting for continual learning, where data arrive in a non-i.i.d online stream without explicit task boundaries. Recently, the \emph{Si-Blurry}~\citep{moon2023online} scenario has been proposed to better reflect the real world. Several methods have been proposed under this scenario. MVP~\citep{moon2023online} introduces a prompt tuning approach based on a pool of learnable prompts. It selects the most appropriate prompt via a query-key matching mechanism and applies instance-wise logit masking. To address class imbalance, MVP further proposes feature scaling and a gradient similarity-based focal loss. Online-LoRA~\citep{wei2024online} extends the LoRA to the continual learning setting by incrementally adding low-rank adaptations to the query and value projection matrices in the self-attention blocks. It proposes an online parameter importance estimation mechanism to merge new LoRA without interference. MISA~\citep{kang2025advancing} argues that prompts, like pretrained models, can benefit from an initial learning session. In addition to this pretraining phase, MISA introduces batch-level masking to prevent forgetting.

\section{Conclusion}
\label{conclusion}

In this paper, we revisit the role of prompt selection in task-free online continual learning. While prompt selection has become dominant, our in-depth analysis reveals that existing methods often fail to consistently associate prompts with input semantics. Motivated by these observations, we propose SinglePrompt, a simple yet effective framework that eliminates the need for prompt selection by reusing a single prompt across all inputs. Built on a minimal architecture with prefix tuning, cosine similarity-based logit calculation and minibatch logit masking, our method achieves strong performance with significantly fewer learnable parameters.
We believe our findings offer a new perspective on prompt-based adaptation in continual learning and can serve as a foundation for future research in this direction.

\section*{Acknowledgment}
This work was partly supported by the grants from Institute of Information \& Communications Technology Planning \& Evaluation (IITP), funded by the Korean government (MSIT; Ministry of Science and ICT): No. RS-2019-II190421, No. IITP-2026-RS-2020-II201821, No. IITP-2026-RS-2024-00437633, and No. RS-2025-25442569.

{
    \small
    \bibliographystyle{ieeenat_fullname}
    \bibliography{main}
}

\clearpage
\setcounter{page}{1}
\maketitlesupplementary

\appendix
\label{sec:appendix}

\section{Evaluation Metrics}
In this section, we provide a detailed description of the evaluation metrics. Let $a_{t, i}$ denote the classification accuracy evaluated on the $i$-th task after training up to the $t$-th task, given a total of $T$ tasks. After completing training on all $T$ tasks, the last accuracy $A_\text{last}$, is defined as:

\begin{equation}
A_\text{last}=\frac{1}{T}\sum_{i=1}^Ta_{T,i}.
\label{eq:A_last}
\end{equation}
\noindent
Also, the forgetting at the last task, $F_\text{last}$, is defined as:
\begin{equation}
F_\text{last}=\frac{1}{T-1}\sum_{i=1}^{T-1}\max_{j\in\{1, \cdots,T-1 \}}(a_{j,i}-a_{T,i}).
\label{eq:F_last}
\end{equation}

The $A_{\text{auc}}$~\citep{koh2022online} is designed to evaluate any time inference accuracy of the model. It is calculated every time the model has trained on $n$ samples. Let $L$ be the total number of evaluation steps, and $a_l$ denotes the accuracy at the $l$-th evaluation step. We define:
\begin{equation}
A_\text{AUC}=\frac{1}{L}\sum_{l=1}^L a_l.
\label{eq:A_auc}
\end{equation}
Note that each evaluation is performed on all the classes the model has learned so far.

\begin{figure}[h]
    \centering
    \includegraphics[width=0.85\linewidth]{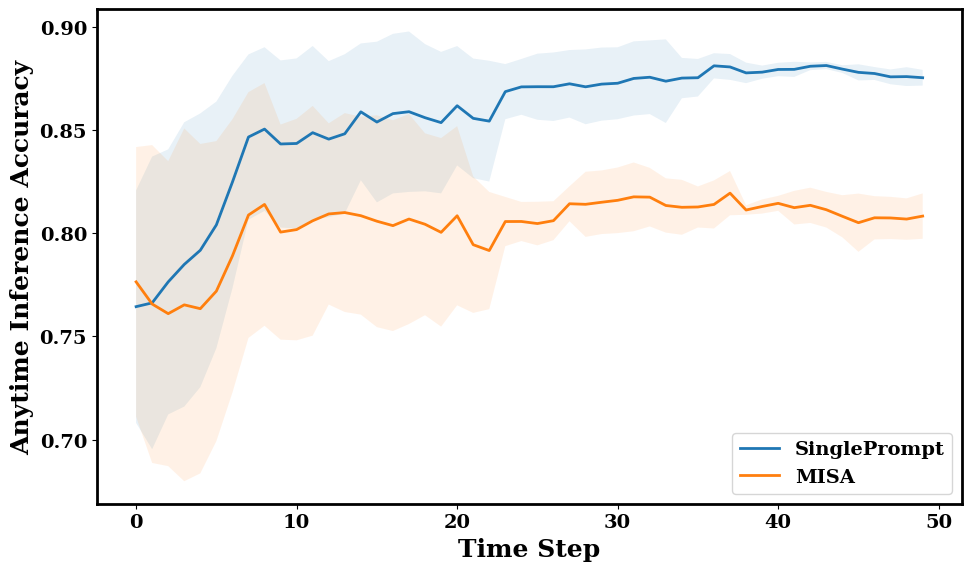}
    \vspace{-0.1in}
    \caption{Anytime inference accuracy curves of SinglePrompt and MISA~\citep{kang2025advancing} on CIFAR100~\citep{krizhevsky2009learning}. Accuracy is measured every 1,000 training samples, and the curves visualize the mean and standard deviation over five random seeds.}
    \label{fig:auc_plot}
\end{figure}

\section{Si-Blurry Scenario}
\begin{figure}[t]
    \centering
    \includegraphics[width=0.85\linewidth]{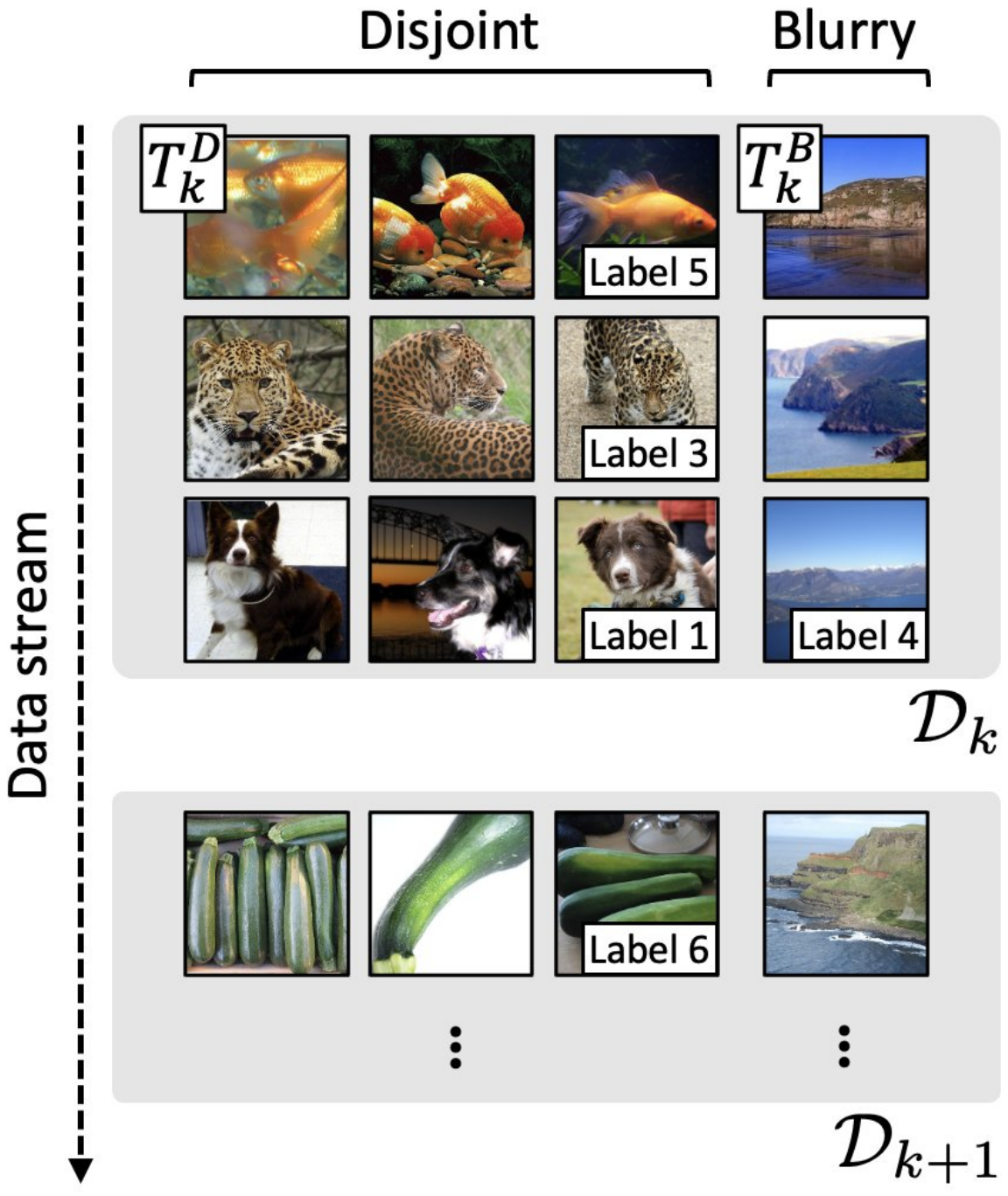}
    \vspace{-0.3in}
    \caption{Visualization of the Si-blurry scenario. The dataset for the $k$-th task, $\mathcal{D}_k$, consists of the disjoint dataset $T_k^D$ and the blurry dataset $T_k^B$.}
    \label{fig:si_blurry}
\end{figure}

To evaluate the effectiveness of our proposed framework in task-free online continual learning, we conducted experiments under the Stochastic incremental-Blurry (Si-Blurry)
scenario~\citep{moon2023online} (see Figure \ref{fig:si_blurry}). A total of $C$ classes are randomly partitioned into disjoint and blurry classes according to a predefined disjoint class ratio. Each class subset is then randomly divided into $T$ tasks. For the $k$-th task, we denote the disjoint and blurry datasets as $T_k^D$ and $T_k^B$, respectively. Among the blurry class samples, a proportion determined by the blurry sample ratio is randomly shuffled across tasks, regardless of task boundaries. This process causes the boundaries between tasks to become ambiguous. As a result, the dataset for the $k$-th task $\mathcal{D}_k$ consists of both $T_k^D$ and $T_k^B$.

\begin{algorithm}[t] 
\caption{\textbf{SinglePrompt}: Pytorch-like Pseudocode}
\label{alg:algorithm}
\textbf{Input}: Input image $x$, label $y$ \\
\textbf{Parameter}: Prompt injected layers $K$, self-attention blocks depth $L$, temperature $\tau$, mask vector $m$ \\
\textbf{Output}: Cross-entropy loss
\begin{algorithmic}[1] 
\STATE $h_0 = f_0(x)$  \quad // $f_0$: input embedding layer
\FOR{$i=1$ \textbf{to} $L$}
    \IF{$i\leq K$}
        \STATE $h_i=f_i\big(h_{i-1};p^k_i,~p_i^v\big)$ \quad // $f_i$: $i$-th self-attention block, $p^k_i,~p_i^v$: prompt for block $i$ (attention operation via equation~\eqref{eq:attn_prefix})
    \ELSE
        \STATE $h_i=f_i(h_{i-1})$ \quad // no prefix tuning (attention operation via equation~\eqref{eq:attn_ori})
    \ENDIF
\ENDFOR
\STATE $\texttt{logits}=\frac{\mathbf{g} \cdot \mathbf{c}}{\|\mathbf{g}\| \|\mathbf{c}\|} \cdot \frac{1}{\tau}+ m$ \quad // $\mathbf{g}$: encoder output feature, $\mathbf{c}$: class prototypes
\STATE $\texttt{loss}=\texttt{CrossEntropy}(\texttt{logits},~y)$
\STATE \textbf{return} $\texttt{loss}$
\end{algorithmic}
\end{algorithm}

\section{Experimental Details}
Pseudo-code for the proposed \textbf{SinglePrompt} is provided in Algorithm~\ref{alg:algorithm}.

\begin{figure}[t]
     \centering
     \begin{subfigure}{0.49\linewidth}
         \centering
         \includegraphics[width=\linewidth]{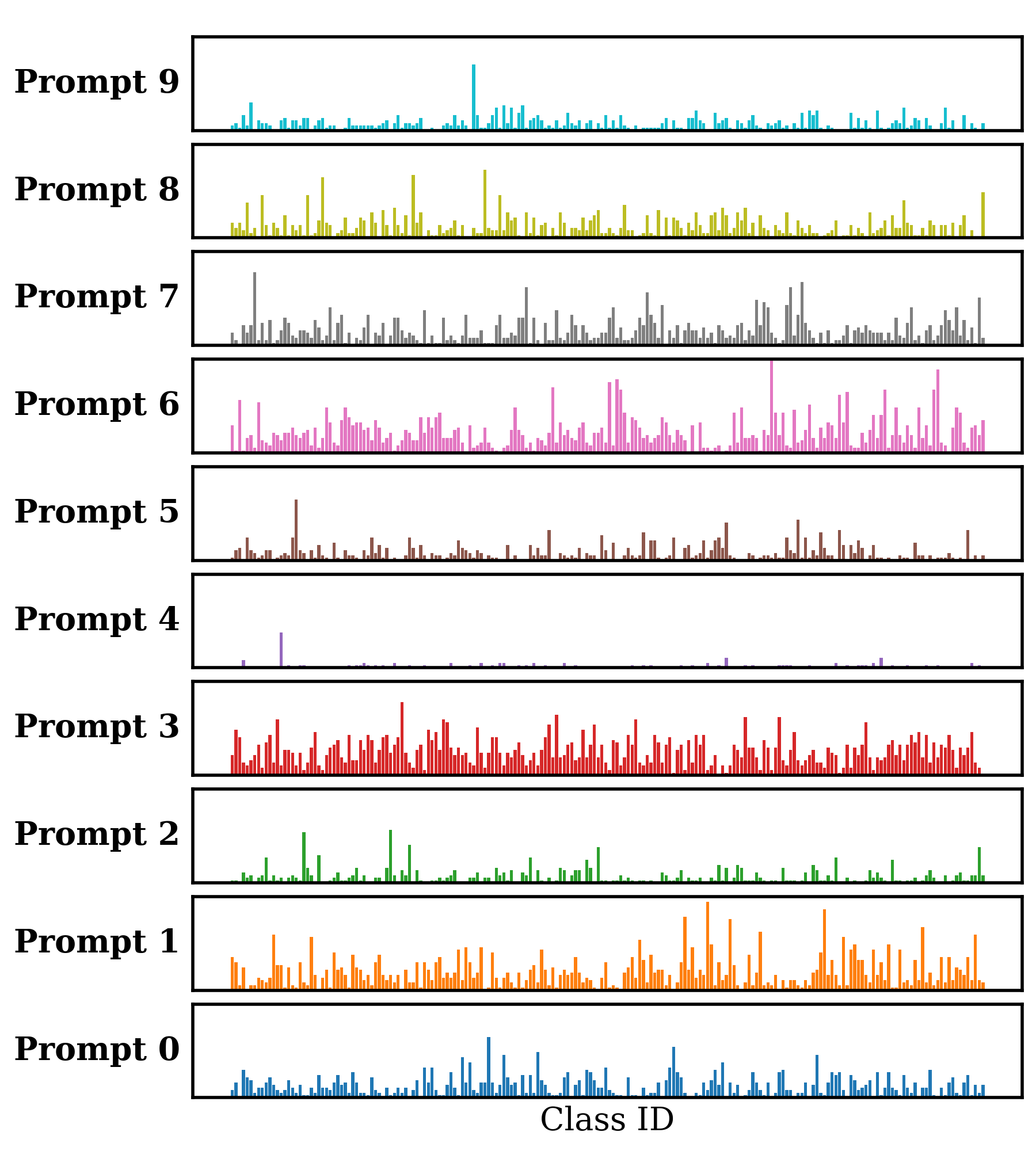}
         \caption{MVP on TinyImageNet}
         \label{fig:tiny_mvp}
     \end{subfigure}
     \hfill
     \begin{subfigure}{0.49\linewidth}
         \centering
         \includegraphics[width=\linewidth]{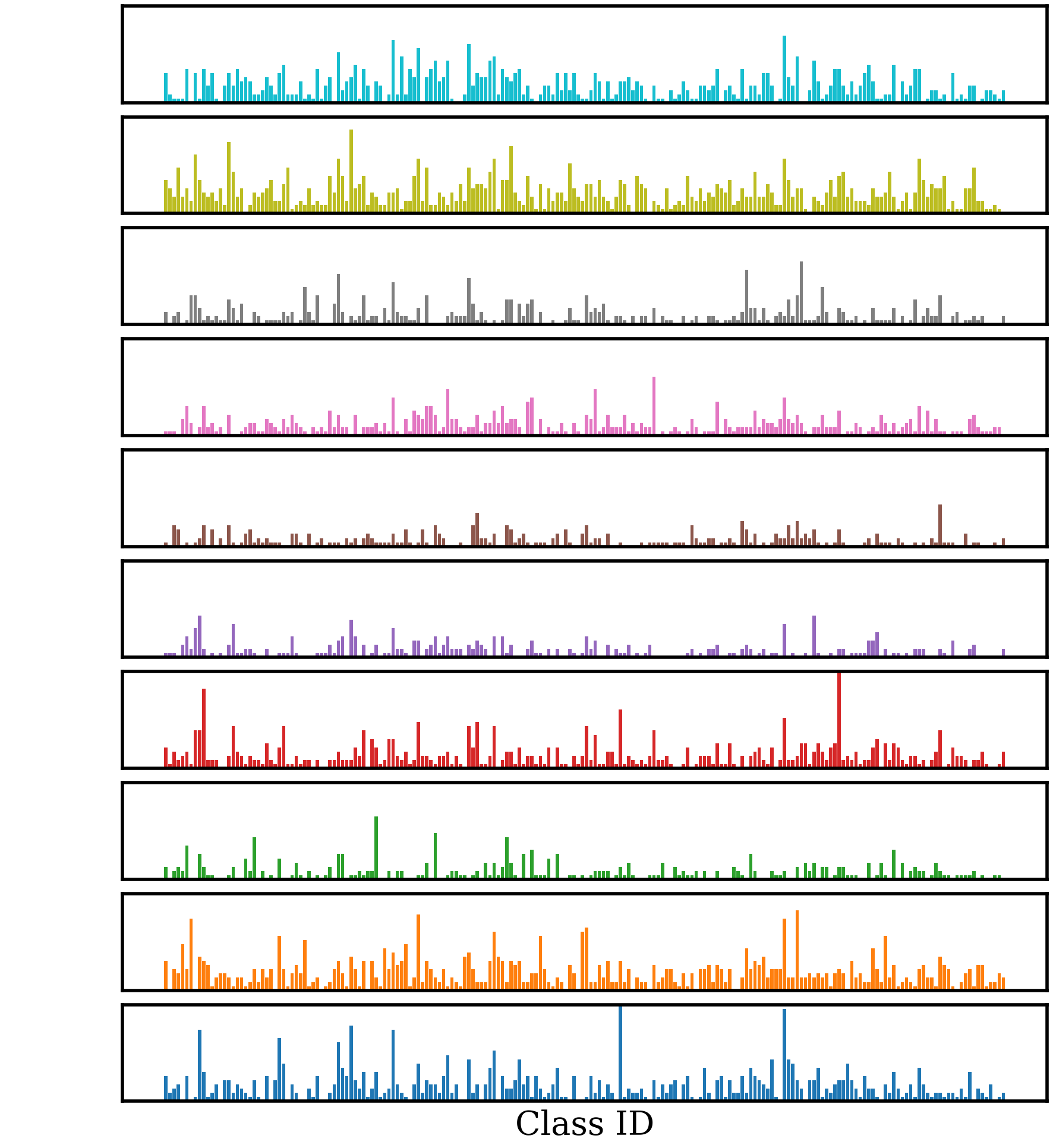}
         \caption{MVP on ImageNet-R}
         \label{fig:imgr_mvp}
     \end{subfigure}
     \vspace{0.5em}
     \centering
     \begin{subfigure}{0.49\linewidth}
         \centering
         \includegraphics[width=\linewidth]{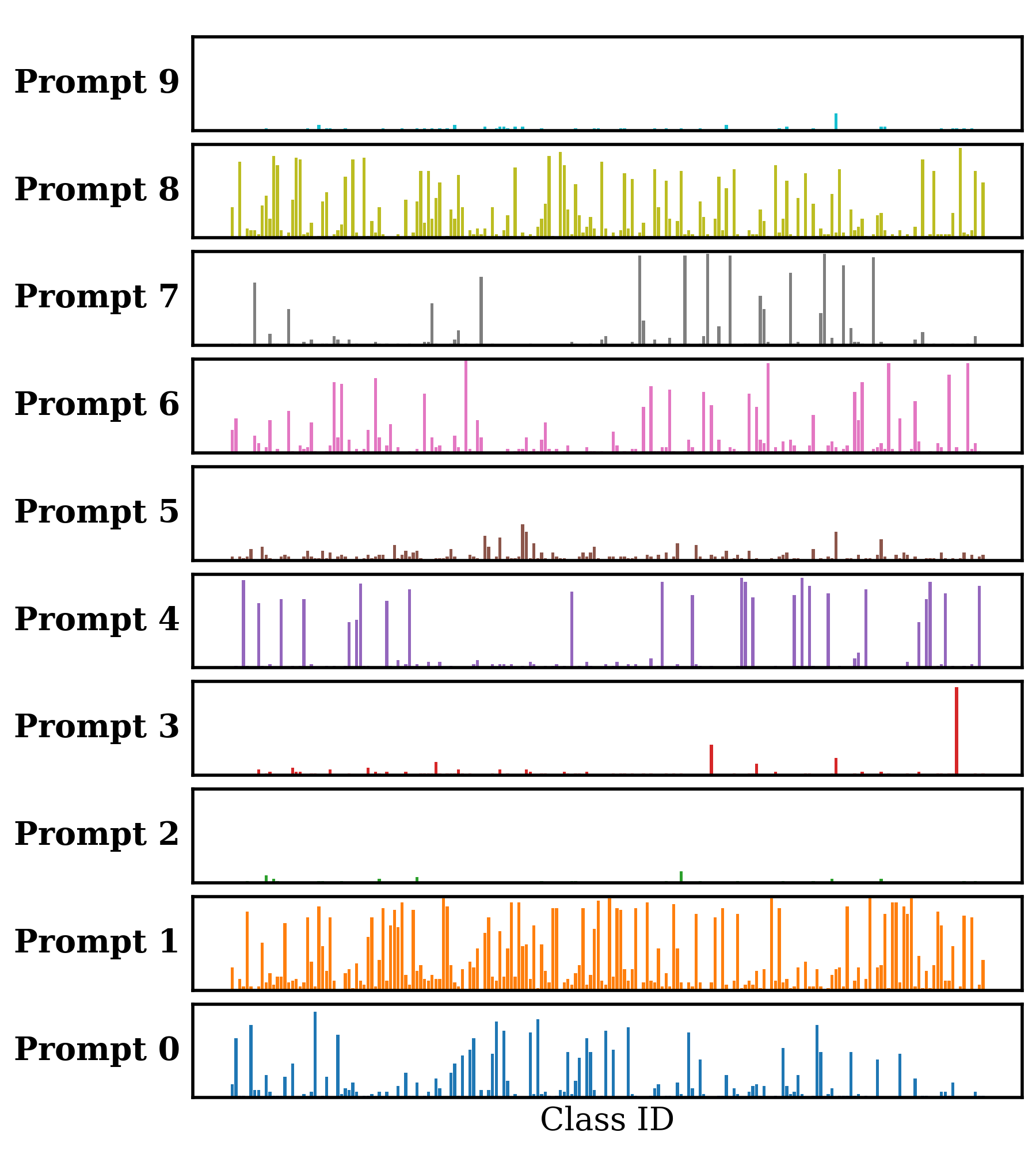}
         \caption{MISA on TinyImageNet}
         \label{fig:tiny_misa}
     \end{subfigure}
     \hfill
     \begin{subfigure}{0.49\linewidth}
         \centering
         \includegraphics[width=\linewidth]{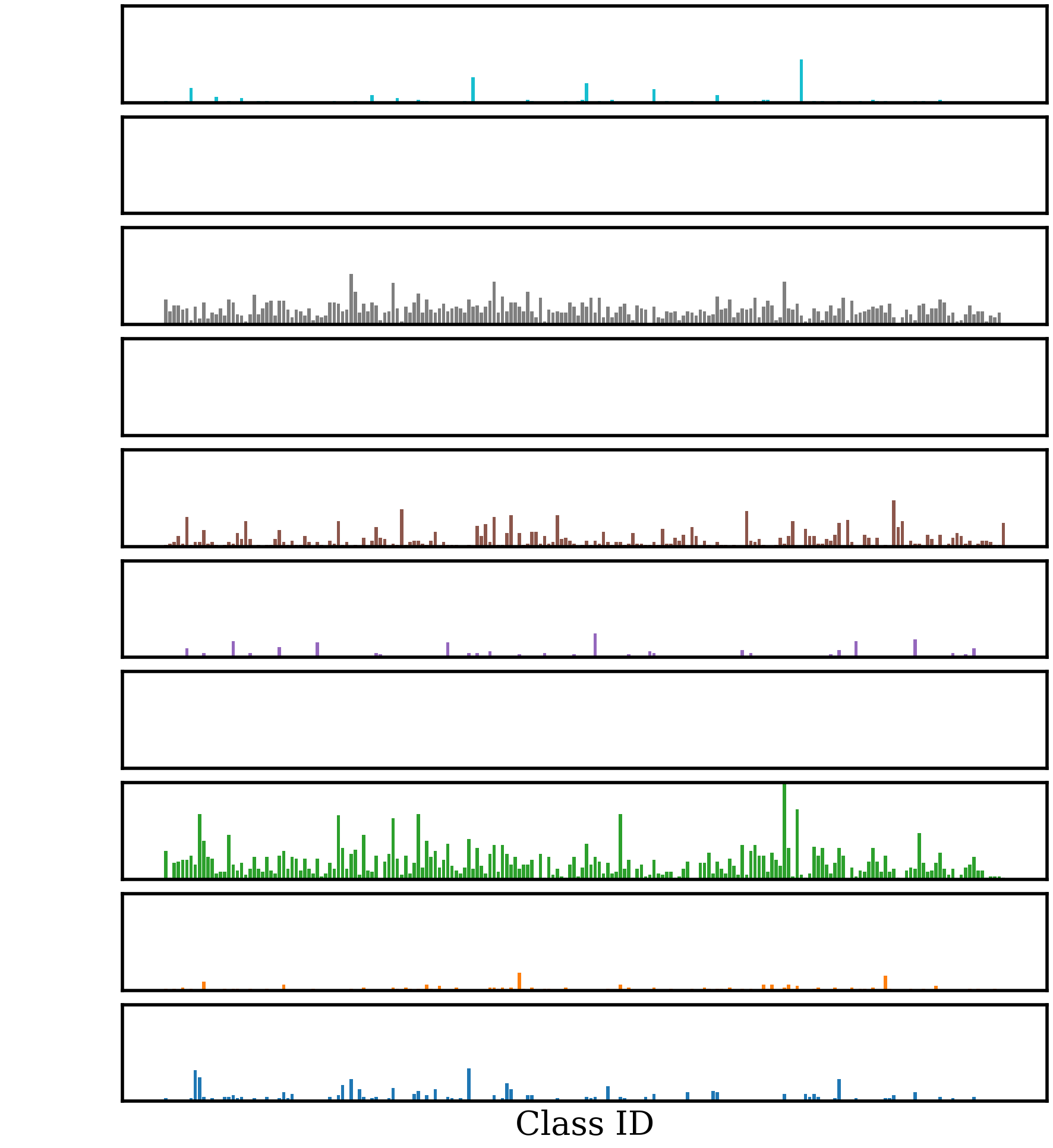}
         \caption{MISA on ImageNet-R}
         \label{fig:imgr_misa}
     \end{subfigure}
     \caption{Prompt selection failures on additional datasets, showing that the observed failure patterns generalize beyond CIFAR-100.}
     \label{fig:fail_other}
\end{figure}

\begin{figure}[t]
     \centering
     \begin{subfigure}{0.49\linewidth}
         \centering
         \includegraphics[width=\linewidth]{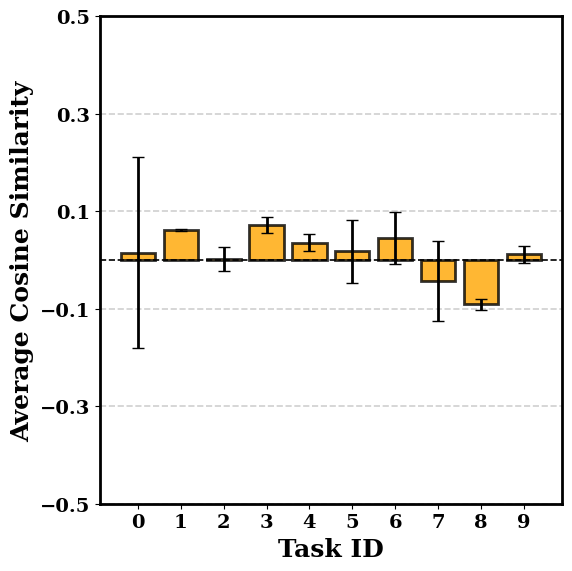}
         \caption{Offline Setup}
         \label{fig:conv_off}
     \end{subfigure}
     \hfill
     \begin{subfigure}{0.49\linewidth}
         \centering
         \includegraphics[width=\linewidth]{figures/convprompt_cnt.png}
         \caption{Online Setup}
         \label{fig:conv_on}
     \end{subfigure}
     \caption{Failure of ConvPrompt~\citep{roy_2024_CVPR} selection in task-based continual learning on CIFAR100~\citep{krizhevsky2009learning}. The x-axis represents the task ID of input sample and the y-axis indicates the average cosine similarity between each task's samples and their assigned keys. (a) Result in the offline continual learning setting, where class information of upcoming tasks is available. Task similarity is computed using class descriptors, which determines the number of prompts allocated to each task (22 prompts in total). (b) Result in the online continual learning setting, where future task classes are unknown. Therefore, the number of prompts assigned to each task is fixed to 5, resulting in 50 prompts in total. In both cases, the cosine similarity is expected to be high for all tasks but remains close to zero. Results are shown for the 7-th layer. Note that consistent results are observed across other layers as well. }
     \label{fig:conv_onoff}
\end{figure}

\paragraph{In-depth Analysis of Prompt Selection}
In Section~\ref{sec:indepth_analysis}, we present five types of prompt selection failures in continual learning. For L2P~\citep{wang2022learning}, DualPrompt~\citep{wang2022dualprompt}, MVP~\citep{moon2023online}, and MISA~\citep{kang2025advancing}, we follow the original experimental setups described in their respective papers. While the main paper reports results on CIFAR100, we extend the analysis to additional datasets and observe that similar prompt selection failures consistently occur, as illustrated by the results of MVP and MISA in Figure~\ref{fig:fail_other}.
For ConvPrompt~\citep{roy_2024_CVPR}, we analyze the effectiveness of its prompt selection mechanism under both online and offline continual learning settings. In the offline setting, where the method is originally introduced, task boundaries are explicitly provided, and the similarity between the upcoming task and previously learned tasks is measured using GPT-3 generated class descriptors. Based on this task similarity, the number of additional prompts is dynamically determined, resulting in the final size of $P=22$. In the online setting, however, the upcoming class information is not available. Consequently, the number of prompts allocated to each task is fixed to 5, leading to the final size of $P=50$. ConvPrompt performs prompt selection through a soft routing mechanism, where the cosine similarity between the class token from the previous layer and a set of learned keys is used as routing weights. The prompts associated with each key are then combined into a weighted sum according to these similarity scores to produce the final prompt for prefix tuning. If the prompt selection mechanism operates as intended, each input sample should exhibit high similarity to the prompt key corresponding to its task. To examine this, we compute and compare the average similarity between input samples and the keys associated with each task's prompts. As shown in Figure~\ref{fig:conv_onoff}, the average cosine similarity across all tasks is expected to be high, yet is observed to be close to zero. This holds for both the online (Figure~\ref{fig:conv_on}) and offline (Figure~\ref{fig:conv_off}) settings, indicating that soft routing mechanism fails in task-based continual learning. Note that the result in the main paper is reported under the online setting.

\paragraph{Data preprocessing.}
We use CIFAR100~\citep{krizhevsky2009learning}, Tiny ImageNet~\citep{le2015tiny} and ImageNet-R~\citep{hendrycks2021many} following prior works. Since we use the ViT-B/16, all input images are resized to 224×224. For each dataset, additional preprocessing was applied in the same manner as in previous studies~\cite{moon2023online, kang2025advancing}.

\paragraph{Computing Infrastructure.}
All experiments were conducted on a machine equipped with an NVIDIA GeForce RTX 4090 GPU, an Intel Xeon Gold 6526Y CPU, and 512 GB of RAM. The code was implemented using PyTorch 2.4.1+cu121, with CUDA 12.1 and cuDNN 9.0.1. Computational statistics are reported in Table~\ref{tab:computational_stat}.

\begin{table}[t]
    \centering
\scalebox{0.9}{%
    \begin{tabular}{lccc}
    \toprule
    Method & \#Params & FLOPs & Total time (s) \\ 
    \midrule
     MVP\citep{moon2023online}
     & $562K$ & $34.21G$  & 2731.43 \\
     MISA\citep{kang2025advancing}
     & $576K$ & $35.67G$ & 3405.24 \\
     Online-LoRA\citep{wei2024online}
     & $524K$ & $17.64G$ & 1764.47 \\
    \cellcolor{gray!15}SinglePrompt& \cellcolor{gray!15}$230K$ & \cellcolor{gray!15}$17.61G$ & \cellcolor{gray!15}1589.74 \\
    \bottomrule
    \end{tabular}}
    \caption{Computational statistics for ours on CIFAR100~\citep{krizhevsky2009learning} without buffer. We compare our method with MISA~\citep{kang2025advancing}, MVP~\citep{moon2023online} and online-LoRA~\citep{wei2024online}. FLOPs refer to forward pass computations only, and total time is averaged over five runs with different random seeds. \#Params refers to the number of learnable parameters. All methods use the same ViT-B/16 architecture to ensure a fair comparison.}
    \label{tab:computational_stat}
\end{table}

\end{document}